\documentclass[10pt,twocolumn,letterpaper]{article}

\usepackage{wacv}
\usepackage{times}
\usepackage{epsfig}
\usepackage{graphicx}
\usepackage{amsmath}
\usepackage{amssymb}
\usepackage{booktabs}

\wacvfinalcopy

\ifwacvfinal
\usepackage[breaklinks=true,bookmarks=false]{hyperref}
\else
\usepackage[pagebackref=true,breaklinks=true,colorlinks,bookmarks=false]{hyperref}
\fi

\begin{document}
	
	%%%%%%%%% TITLE
	\title{\LaTeX\ Author Guidelines for WACV Proceedings}
	\title{Combining Photogrammetric Computer Vision and Semantic Segmentation for Fine-grained Understanding of Coral Reef Growth under Climate Change}
	
	\author{
	Jiageng Zhong\\
	WHU\\
	%China\\
	{\tt\small zhongjiageng@whu.edu.cn}
	\and
	Ming Li\\
	ETH Zurich, WHU\\
	%Switzerland\\
	{\tt\small mingli39@ethz.ch}
	\and
	Hanqi Zhang\\
	WHU\\
	%China\\
	{\tt\small hqzhang@whu.edu.cn}
	\and
	Jiangying Qin\\
	WHU\\
	%China\\
	{\tt\small jy\_qin@whu.edu.cn}
}

\maketitle
\thispagestyle{empty}

%%%%%%%%% ABSTRACT
\begin{abstract}
	Corals are the primary habitat-building life-form on reefs that support a quarter of the species in the ocean. A coral reef ecosystem usually consists of reefs, each of which is like a tall building in any city. These reef-building corals secrete hard calcareous exoskeletons that give them structural rigidity, and are also a prerequisite for our accurate 3D modeling and semantic mapping using advanced photogrammetric computer vision and machine learning. Underwater videography as a modern underwater remote sensing tool is a high-resolution coral habitat survey and mapping technique. In this paper, detailed 3D mesh models, digital surface models and orthophotos of the coral habitat are generated from the collected coral images and underwater control points. Meanwhile, a novel pixel-wise semantic segmentation approach of orthophotos is performed by advanced deep learning. Finally, the semantic map is mapped into 3D space. For the first time, 3D fine-grained semantic modeling and rugosity evaluation of coral reefs have been completed at millimeter (mm) accuracy. This provides a new and powerful method for understanding the processes and characteristics of coral reef change at high spatial and temporal resolution under climate change.
\end{abstract}

%%%%%%%%% BODY TEXT
\section{Introduction}

Coral reefs, as the most prominent and representative ecosystems in clear warm tropical and subtropical oceans, possess astonishing biodiversity and structural complexity, as well as extremely high primary productivity~\cite{mellin2022safeguarding1, bowen2013origins2}. The coral reef ecosystem is an essential natural resource - for humanity and all marine ecosystems. But while resources are strictly finite, human demands are not. Ecologists see them as the upper limit of what marine ecosystems can evolve to, always comparing them to tropical rainforests on land. The coral reef biome is the basic structure of the coral reef ecosystem, which is composed of scleractinian corals, reef-building algae, and a rich variety of reef-dwelling animals and plants. Coral reefs are facing increasing stress from global climate change combined with local stresses from sedimentation, resource extraction, over-fishing and other sources of land-based pollution~\cite{hughes2017global3, carlson2019land4}. Today, this has led to events such as coral degradation, increased ocean flooding, and loss of coral reef flora and fauna. 14\% of corals have been lost over the past decade, and if humans cannot control global warming to within 1.5° in 2050, 70-90\% of corals will be lost~\cite{TP-toolbox-web5, allan2021ipcc6, portner2022climate7}. These factors that affect coral reef health and sustainability are diverse. How much pressure does the coral reef endure and how can we minimize it? Both public and private organizations can play a vital role in protecting our coral reefs, and understanding is the first step. Therefore, in order to explore the fascinating underwater world of coral reefs, how to use advanced technology to map, monitor and model coral reef habitats is the key to our better understanding, protection and preservation of coral reefs. Considering the potential for these stressors to increase the rate of coral degradation to dire levels, ecologists and administrators worldwide are shifting their focus from prevention to identifying key mechanisms controlling reef resiliency and determining where coral reefs will be the most viable in the future. So, identifying habitats with the highest survival potential is critical to coral reef conservation and finding these refuges requires long-term observations of coral reefs to identify these potential habitats.

Manual local in situ underwater surveys have long been the primary method for assessing coral distributions and growth health, but it requires a substantial investment of time during a field diver survey, this limits the spatial and temporal scales over which ecological surveys can be conducted~\cite{zhang2022deep8}. In the past decade, new methods for mapping benthic habitats have been developed using satellite and aerial photogrammetry and remote sensing. These methods can quickly provide information on large-scale coral monitoring projects and enable the identification of different benthic function types on coral reefs, but they cannot provide detailed observations of coral complex structures due to the water surface effect~\cite{zhang2022deep8, candela2021using9}. The emergence of underwater photogrammetry and unmanned underwater vehicles have greatly improved the data collection method of in-water surveys, providing a millimeter-level high-resolution monitoring capability for coral reef observation and allowing us to observe even individual corals~\cite{johnson2017high10,guo2016accuracy1armin,nocerino2020coral2armin}. However, it also brings data processing bottlenecks and technical complexity that are difficult to be processed manually by traditional means~\cite{beijbom2012automated11}. Fortunately, recent rapid advances in the repeatability and availability of photogrammetric computer vision and machine learning automated tools have gradually been removing some of these barriers~\cite{hopkinson2020automated12}. This has the potential to address the long-standing challenges of monitoring rapid changes in coral reefs with high spatiotemporal resolution and reproducibility, and help us understand coral reef vulnerability and resilience in the face of global and local stress, especially global climate change~\cite{asner2020high13, neyer2018monitoring3armin}.

To implement fine-grained understanding of coral reef growth variations, and to serve the long-term monitoring of coral reefs, we propose a new approach from a new perspective, which combines the technical advantages of photogrammetric computer vision and machine learning. The advanced photogrammetric computer vision technology is applied first to reconstruct the coral reef in 3D, and outputs products including high-resolution and high-accuracy underwater digital surface models (DSM), orthophotos and 3D meshes. A novel neural network for semantic segmentation of underwater coral images using orthophotos and DSMs is then developed. Finally, based on these results from different periods, the coral reef growth variations can be intelligently analyzed in a three-dimensional fashion which will help marine ecologists a lot.

\section{Related works}

{\bf Photogrammetric Computer Vision:} In photogrammetric computer vision, many approaches are providing new automated image processing tools to generate high-resolution 3D models for capturing the spatial structural complexity of coral reef ecosystems~\cite{edwards2017large14,burns2016assessing15}. However, historically coral cover has been a biological indicator of resilience that the scientific community has primarily focused on. Coral reefs have often been studied as a two-dimensional system at this time. It is clear that the percent cover indicator is not sufficient to reflect the structural complexity of coral reefs, which constitute the diversity and importance of the ecosystem, and structurally complex coral reefs should be studied most accurately in three-dimensional space~\cite{friedlander1998habitat16}. The 3D rugosity index is a useful metric to assess the structural complexity of reefs at small to medium spatial scales linked to the high diversity of organisms on reefs~\cite{asner2020high13,graham2013importance17}. The Vector Ruggedness Measure is a widely used surface roughness metric that incorporates the variation of slope and aspect into a single measurement~\cite{sappington2007quantifying18,price2019using19}. Over the past decade, Simultaneous Localization and Mapping (SLAM) or Structure-from-Motion (SfM) techniques were used to process the underwater coral images to obtain 3D reconstructions of coral reefs, and employed to better understand the spatial clustering of species, the effects of disturbance on coral reef complexity and community structure~\cite{johnson2010generation20}. % ,burns2015integrating21,storlazzi2016end22 

{\bf Coral Image Segmentation:} The advances in machine learning have resulted in significant progress in image segmentation and object classification. In machine learning, more recently progress has been made in training automated classifiers to classify or segment underwater images to count the species abundance in marine ecosystems~\cite{hopkinson2020automated12}. For example, traditional machine learning methods (SVMs, k-nearest neighbor, etc.) and deep learning are used to estimate coral percent cover: the percentage of surface occupied in the surveyed area by a given taxa or substrate when viewed from overhead~\cite{hopkinson2020automated12}. These classifiers can obtain promising results for most classes of coral reefs based on local texture and color features. The best performing traditional machine learning classifiers (SVMs) have high classification accuracy on abundant classes, reaching around 80\%, but their accuracy drops drastically on less common classes and can only resolve coral classes to the genus or functional group level~\cite{beijbom2015towards23}. Recently, Convolutional Neural Network (CNN)-based classifiers have shown promise over SVM-based classifiers for segmenting coral images~\cite{alonso2017coral24}. Patch-based CNN approaches are a major kind of deep learning methods for semantic segmentation~\cite{king2018comparison25}. However, it is important to note that the granularity of classification is not fine enough with a patch-based CNN model since it gives a single class label for an entire patch within an image. \cite{king2018comparison25,wang2018understanding26,long2015fully} present the Fully Convolutional Neural Network (FCNN) model, which represents modifications of traditional CNNs to output full semantic segmentation of input images at the pixel level. Unlike patch-based CNN models, FCNN models have no limitations on localization accuracy, and they provide a classification for each individual pixel within an image. Furthermore, advanced classification techniques have been applied to classify 3D coral reef reconstructions, which can provide new insights into the spatial distribution relationships among coral taxa and offer more realistic representations of the biomass of organisms in coral reef systems compared to two-dimensional metrics such as coral percent cover~\cite{hopkinson2020automated12,edwards2017large14}. However, they did not consider the use of underwater control points in the classification of 3D coral reconstruction, and did not use pixel-wise semantic segmentation results to analyze changes intraspecific and interspecific of corals, which led to the inability to accurately monitor changes of coral reefs with high spatial accuracy at different times. Of course, the previous works effectively mitigate a fundamental problem limiting the study of coral reef ecology: the difficulty of generating accurate and reproducible semantic maps of coral reef habitats.

\section{Materials and methods}

\begin{figure}[t]
	\centering
	\includegraphics[scale=0.27]{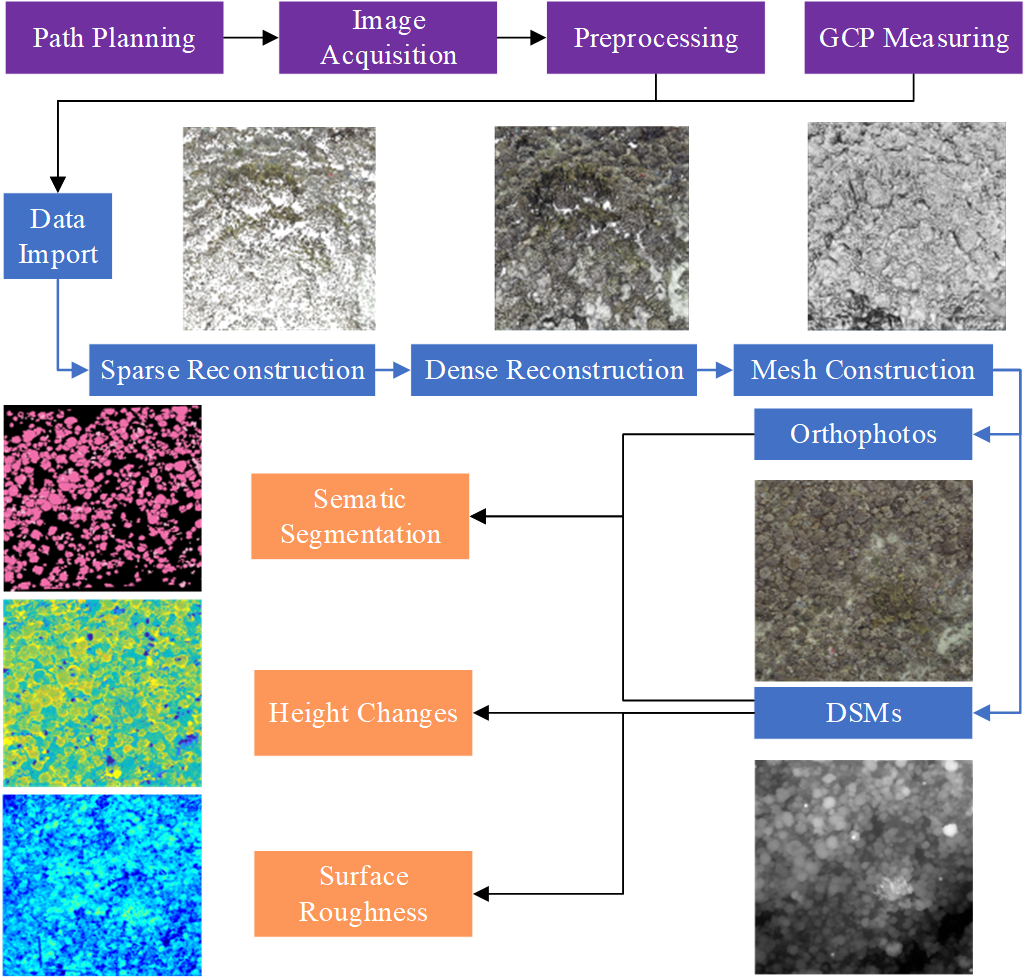}
	\caption{A workflow for the fine-grained understanding of coral reef growth variations.}
	\label{fig:1}
\end{figure}

Our approach for fine-grained understanding of coral reef growth variations is shown in Figure \ref{fig:1}. First, prepare data, including preprocessed underwater images and measurements related to ground control points (GCPs). Next, import data to the photogrammetric computer vision program, and obtain a set of products, i.e., sparse point cloud, dense point cloud, mesh model, orthophoto and DSM. Then orthophotos and DSMs are used to classify coral reefs through deep learning methods, and DSMs can also be used to calculate the reef height changes and surface roughness. Finally, we get a variation monitoring measurements about coral reefs from the above different processing phases, which are provided to coral ecologists for further analyses from different perspectives. Our source code can be found at \href{https://github.com/Atypical-Programmer/Coral-3D-Analysis-Toolbox}{https://github.com/Atypical-Programmer/Coral-3D-Analysis-Toolbox}.

\subsection{Data acquisition and preprocessing}

\begin{figure}[t]
	\centering
	\includegraphics[scale=0.28]{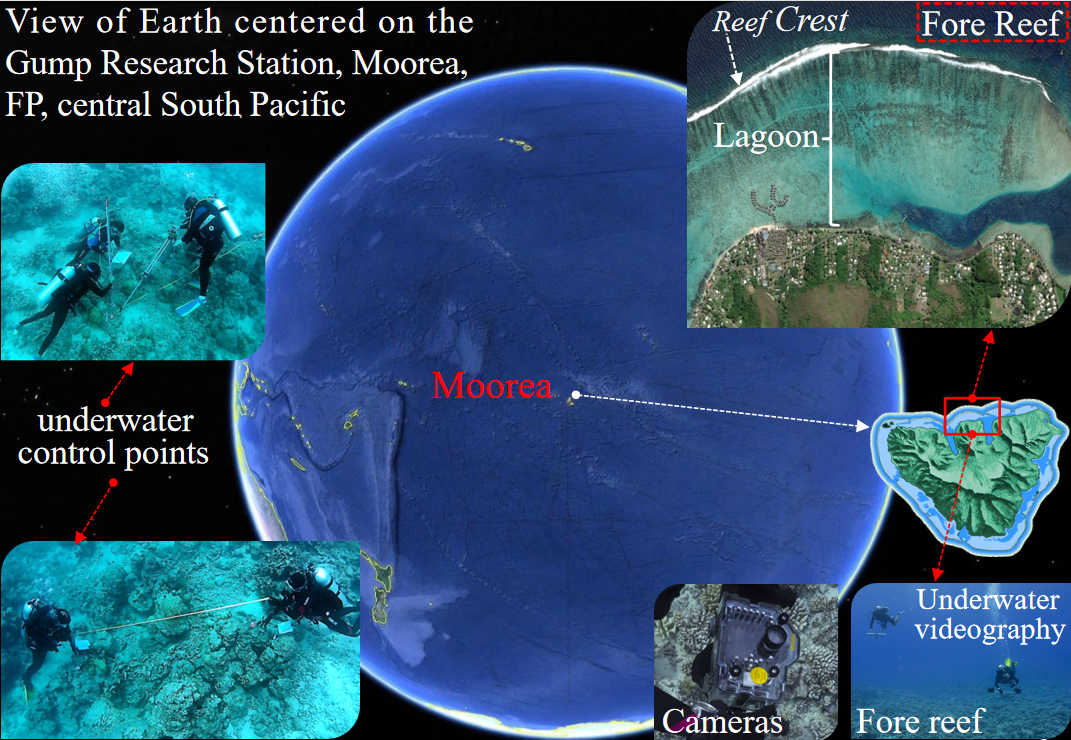}
	\caption{Underwater coral data acquisition locations and some key field work scenarios.}
	\label{fig:2}
\end{figure}

This study is supported by Moorea Island Digital Ecosystem Avatar (IDEA) project, and Moorea is a volcanic island in French Polynesia. It is an atoll with about 10 enclosed coral reefs surrounding the entire island. Richard B. Gump South Pacific Research Station on the west coast of Cook's Bay which is home to the Moorea Coral Reef Long Term Ecological Research Site (MCR LTER), part of a network established by the U.S. National Science Foundation in 1980 to support research on long-term ecological phenomena. This is an excellent location for coral monitoring in the South Pacific, and has a wide variety of coral observational data. The data for this study came from underwater remote sensing imagery collected in cooperation with the Gump Station. This data collection is located at the fore reef in Figure \ref{fig:2}, which is the outside part of a reef seaward of the reef crest (or reef edge) facing the open sea. For the implementation of underwater remote sensing of coral reefs in Moorea Island, an underwater camera system is designed to collect underwater coral images. The height of the camera from the benthos is about 2m. The underwater photos were taken in August 2018 and August 2019, respectively. Due to the complex lighting in underwater imaging, we performed radiometric correction on the acquired images~\cite{neyer2019image42}. 

In the study area, \textit{Pocillopora} coral is an absolute dominant coral species and mainly affected by climate change, which is used as an example to conduct coral semantic segmentation and change monitoring analysis. Furthermore, the orthophoto is used by experts to annotate as training data and ground truth for coral image segmentation. Every pixel in the orthophoto is assigned to one of the following three categories: (1) live \textit{Pocillopora}, (2) dead \textit{Pocillopora}, and (3) background (other corals, sea rods, algae, stones, sand, etc.). The establishment of GCPs on the coral seabed in the study area is crucial to improving the accuracy of photogrammetry, and is also conducive to the monitoring of changes in the three-dimensional spatial structure of corals at different times. In our study area, there are five GCPs, each of which corresponds to a unique pattern that can be automatically recognized and measured by program. GCPs not only define a unique datum over all measurement periods but also are supporting the procedure of self-calibration in bundle adjustment.

\subsection{Generation of orthophotos and DSMs}

The orthophotos and DSMs are produced using a photogrammetric computer vision program based on OpenDroneMap~\cite{OpenDroneMap27} which is a rapidly evolving community-based open-source photogrammetry program for processing and analyzing imagery. OpenDroneMap is a powerful tool to generate high-precision DSM and orthophoto with photography~\cite{rinaldi2021dsm28}, and can run on all major operating systems. In OpenDroneMap, one of the key technologies for reconstruction is SfM. As the model obtained by computer vision-based SfM is initially captured in an arbitrary reference system~\cite{sanz2018accuracy36}, it is hard to quantitatively compare two models built with images from the different times at the same place, which will fail to meet the requirements of further analyses such as height changes at different times. Therefore, it is crucial to perform geo-referencing that transforms this initial arbitrary datum into a predefined coordinate reference system. We used underwater GCP landmarks, and established an underwater geodetic control network using green lasers for leveling and distance measurement equipment (Figure \ref{fig:2}). These landmarks are visible on the images, for which the geospatial position (latitude, longitude and altitude) are known. Every GCP can be observed in one or more images. GCP observations are used to align 3D models and refine reconstructions, turning a free-network model into an aligned model~\cite{yan2017novel47}.

\begin{figure}[t]
	\centering
	\includegraphics[scale=0.13]{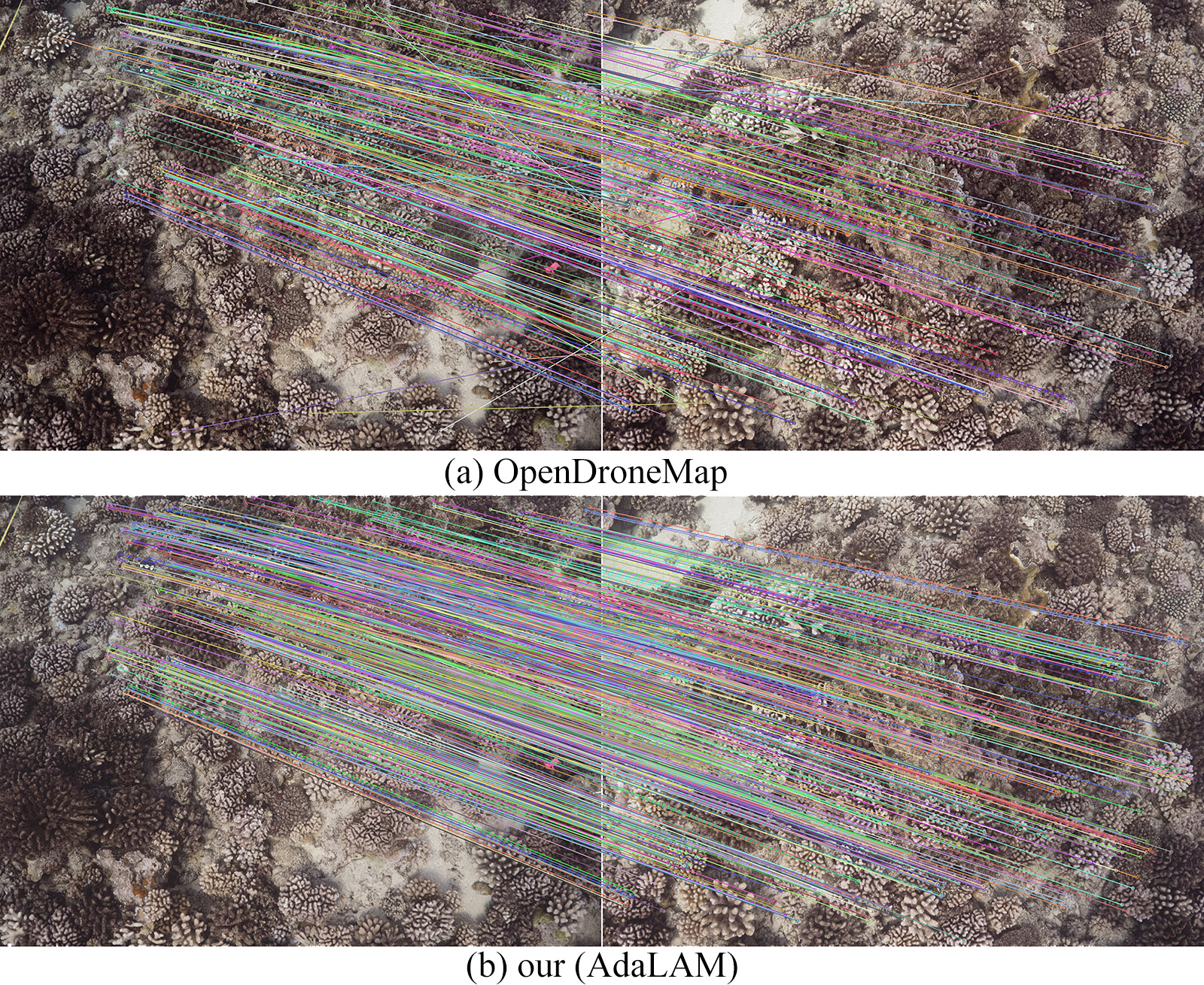}
	\caption{An example for comparison of the matched correspondence of  OpenDroneMap and AdaLAM.}
	\label{fig:4}
\end{figure}

In the feature extraction stage, in order to obtain more accurate Aerial Triangulation (AT) results, we have improved OpenDroneMap. Specifically, we adopt Root SIFT~\cite{arandjelovic2012three31} which can improve the performance of  SIFT~\cite{lowe2004distinctive29} and is obtained by L1 normalizing the SIFT descriptors and taking the square root of each element. In the original OpenDroneMap, Fast Library for Approximate Nearest Neighbors (FLANN)~\cite{muja2009fast30} is applied for feature matching during aligning images. In this study, we use the approach Adaptive Locally-Affine Matching (AdaLAM)~\cite{cavalli2020adalam32} to substitute it, for the reason that it has been proved to be more than competitive to the current state of the art, both in terms of efficiency and effectiveness. As shown in Figure \ref{fig:4}, using AdaLAM for feature matching results in more correct correspondences and fewer wrong correspondences, which makes the reconstruction more robust. From the experiments, AdaLAM can generate nearly 20\% correct correspondences. The photogrammetric computer vision workflow based on tools of OpenDroneMap is shown in the blue flowchart in Figure \ref{fig:1}. After importing the coral images and GCP coordinates, the images are aligned at first with geographical information. By using SfM technology, the positions and orientations of cameras are estimated accurately, and the sparse 3D point cloud of reefs is also generated. Next, through Multi-View Stereo (MVS) 3D reconstructions~\cite{shen2013accurate33} , the point cloud is densified. Then, the dense point cloud is converted into a triangular mesh model using the surface reconstruction method. After building the mesh, orthophoto and DSM can be generated.

\subsection{Coral orthophoto segmentation}

\begin{figure}[t]
	\centering
	\includegraphics[scale=0.165]{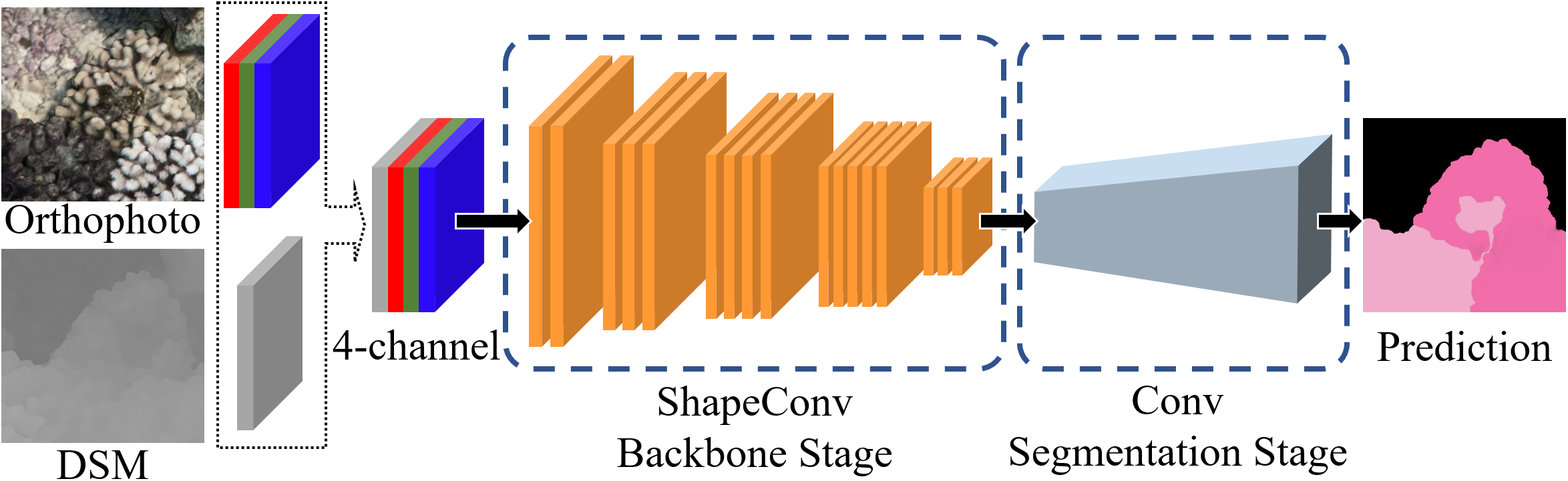}
	\caption{The architecture of MMCS-Net.}
	\label{fig:5}
\end{figure}

In the previous study~\cite{pavoni2019semantic50}, the coral image segmentation is applied on orthophotos which are RGB images. The coral segmentation using orthophotos has many advantages, because orthophotos can incorporate information of the actual metric scale, depth and the geographical coordinates, and reduce labeling time. Some achievements on this have been made over the past years, but there is still vast room for improvement, particularly when faced with challenging factors, such as complicated backgrounds or different lighting conditions in the scenes. It is hard to segment coral precisely only with RGB images. One effective way to overcome these challenges is to employ height, which can provide RGB images with the information contained in the complementary spatial structure of the coral reef. Therefore, we propose the Multi-Modal Coral Segmentation Network (MMCS-Net) which takes patches of the orthophoto and DSM as the input and outputs the coral segmentation result at pixel level.

As DeepLabv3+~\cite{chen2018encoder37} has been proven to achieve the state-of-the-art performance in many previous studies, we decide to build our network based on its architecture. DeepLabv3+ combines the advantages of encoder-decoder architecture and atrous spatial pyramid pooling (ASPP) which can capture rich contextual information from images at various resolutions, and also introduces the idea of depth-wise separable convolution~\cite{howard2017mobilenets41}, which reduces the number of parameters while improving both running speed and classification performance. To leverage the information from DSM, the input is converted from RGB images to RGB images + DSM via channel-wise concatenation. Considering the characteristics of DSM data, we apply the Shape-aware convolutional layer (ShapeConv)~\cite{cao2021shapeconv39} to replace the vanilla convolution layers in the original backbone for processing the height feature. The ShapeConv can effectively leverage the shape information of patches to integrate the RGB and depth cues. Specifically, the height feature is decomposed into a shape-component and a base-component which will cooperate with two learnable weights and be combined by a convolution. It has been proved that the CNNs with ShapeConv can have better performance without introducing any computation and memory increase in the inference phase. The architecture of our segmentation network is shown in Figure \ref{fig:5}.

As for the loss function, the Cross Entropy (CE)~\cite{yi2004automated43} loss is one of the most widely used losses in semantic segmentation. Cross-entropy  is defined as a measure of the difference between two probability distributions for a given random variable or set of events, but it does not consider the labels of neighborhood and it weights both the foreground and background pixels equally~\cite{qin2019basnet46}. To obtain high-quality regional segmentation, we apply a hybrid loss for training: \begin{equation}L=L_{CE}+\mu L_{IoU} \end{equation}
where $L_{CE}$ and ${L_{IoU}}$ denote CE loss and IoU loss, respectively. And $\mu$ is a hyperparameter. We found that $\mu$=0.4 yields good results in practice and we set $\mu$ to this value. IoU is originally proposed for measuring the similarity of two sets, and can be used as the training loss~\cite{mattyus2017deeproadmapper44}. CE loss is pixel-wise, and IoU is a map-level measure. When combining these two losses, we utilize CE to maintain a smooth gradient for all pixels, while using IoU to give more focus on the foreground. 

\subsection{Height changes and surface roughness}

As the georeferenced DSMs from different years have been generated, we subtract them from each other to estimate height changes. The surface height changes over the coral reefs can directly reflect coral growth or degradation during these years and provide reference for the study of the long-term changes in coral reef ecosystems. For example, monitoring disturbance events such as El Niño lead to changes in coral mortality and growth.

The surface roughness is also a critically important measure of reef condition~\cite{asner2020high13}. This metric is usually computed for a square window to quantify the amount of vertical variation in the surface relative to a flat representation of the same region. In marine ecological studies, rugosity is a general parameter because it defines the physical structure of the reef and often correlates positively with the abundance, biomass, and species richness of reef fish~\cite{kuffner2007relationships34}. The Vector Ruggedness Measure (VRM)~\cite{sappington2007quantifying18} is applied here to investigate spatial patterns derived at high spatial resolution. The VRM incorporates variation of slope and aspect into a single measurement. For each cell in a user-defined floating window, a unit vector orthogonal to the cell is decomposed using the 3D location of the cell center along with the slope and aspect. The magnitude is standardized by dividing the number of cells in the neighborhood. Finally, the value is scaled with 0 representing flat and 1 as the most rugged~\cite{asner2020high13}. Specifically, the DSMs are imported into ArcMap 10.7, and VRM is calculated using the Benthic Terrain Modeler tool (BTM)~\cite{walbridge2018unified35}. For the reason that VRM is computed for a square window region, it is inherently dependent upon the scale. In order to assess any measurement scale-dependent relationships, VRM should be calculated with different window sizes for a full-range investigation of scales from individual polyps to colony scales~\cite{price2019using19}. In the past, sonar and other sensors were used to obtain submarine DSM. In this paper, high-precision photogrammetry technology was used to generate fine coral habitat DSM, which is not only low-cost, but also avoids the limitations of using sonar and other equipment in shallow water areas.

\begin{figure*}
	\centering
	\includegraphics[scale=0.32]{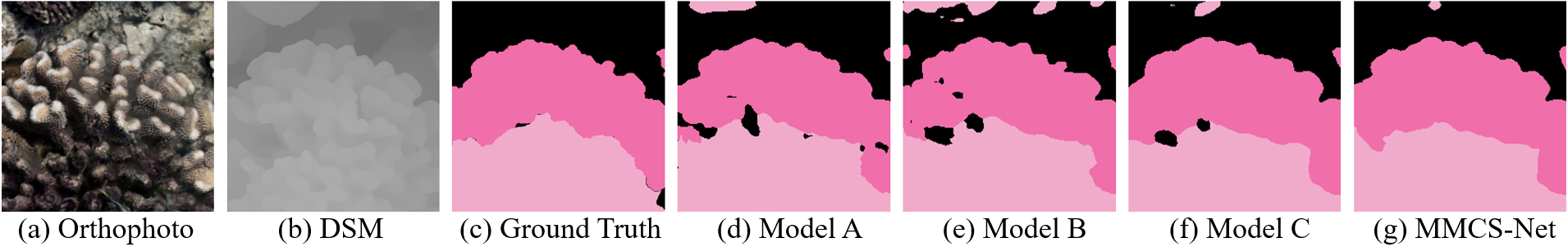}
	\caption{Outputs of different models for a given input.}
	\label{fig:6}
\end{figure*}

\section{Experimental results and discussions}

\subsection{Photogrammetric products}

All coral images collected from the survey are used to generate orthophotos and DSMs. In the photogrammetric process, four GCPs are considered for the geo-referencing process, while the remaining one serves as the check point for accuracy verification. Every GCP can be observed by more than ten images, and their precise 2D coordinates in the image plane are measured. The root mean square errors (RMSE) of check points for the two years are calculated for quality assessment. The horizontal errors are within 4mm, and the total errors are within 5mm. With the use of GCPs, the 3D reconstruction can all be finished with an absolute scale, and maintain accuracy and consistency within the model.

On this foundation, subsequent processes can be carried out. For the sake of fine-grained understanding of coral reefs, the orthophotos and DSMs with a resolution of one millimeter are generated finally. Besides, as for subsequent analyses, there are also intermediate products, including a dense point cloud comprising ten millions of points and a triangular mesh containing ten millions of faces. Consequently, we obtain products with extremely high resolution and high accuracy for fine-grained understanding of coral reef growth variations.

\subsection{Pixel-wise coral segmentation with the trained neural model}

The programs for coral segmentation run on a desktop computer equipped with an NVIDIA GeForce RTX 2080Ti card and 64GB RAM. The networks are implemented with PyTorch deep learning library~\cite{paszke2019pytorch45}. For the preparation of the dataset, the orthophoto is clipped into a set of 448 * 448 tiles through a sliding window with stride 224. For our study area, almost 2000 tiles can be obtained from one orthophoto. Then the dataset is augmented by random translation and rotation. For the validation procedure, a five-fold cross-validation is performed. The data used to train the model are split into five equal parts (folds). Each fold was used once as a validation while the remaining folds were used to train and run the model. The model was run five times and each time the accuracy and loss were calculated. As for the evaluation metrics, the results are reported using Mean Pixel Accuracy (mPA) and Mean Region Intersection over Union (mIoU)~\cite{garcia2017review40}. The metrics from multiple tests of each model are averaged for comparison.

To validate the effectiveness of the added DSM and our improvement, we test four models for ablation experiments (1) Model A: use DeepLabv3+, and only input orthophoto; (2) Model B: use DeepLabv3+, and input orthophoto and DSM; (3) Model C: use DeepLabv3+, replace convolutional layers in the backbone with ShapeConv, and input orthophoto and DSM; (4) Our Model MMCS-Net. Model A, B and C are supervised by CE loss, while MMCS-Net is supervised by hybrid loss.  The results in Table \ref{table:performance} illustrate that the DSM can help to achieve more accurate segmentation, and ShapeConv is more capable to process DSM data to make use of height information. Our method (MMCS-Net) has the best performance due to the above improvement and hybrid loss. Figure \ref{fig:6} illustrates the semantic segmentation results of different models. Model A suffers from poor light caused by occlusions, and Model B performs a little better with the use of height information. By the comparison of Model B with Model C, it can be concluded that ShapeConv can improve the segmentation in edge areas by making better use of structure information than vanilla convolution. Specifically, this is because ShapeConv yields a positive tendency for smoothing neighborhood regions within the same classes~\cite{cao2021shapeconv39}. MMCS-Net achieves the best performance with the application of ShapeConv and a hybrid loss. Due to the supervision of IoU loss, MMCS-Net has a better effect on the whole.

\begin{table} 
	\centering	
	{\small{
			\begin{tabular}{llr}
				\toprule
				Method&mPA&mIoU \\
				\midrule					
				Model A (DeepLabv3+)&89.9\%&80.5\%\\
				Model B&90.8\%&82.1\%\\
				Model C&91.6\%&83.5\%\\
				{\bf MMCS-Net}&{\bf 92.2\%}&{\bf 84.7\%}\\				
				\bottomrule
			\end{tabular}
	}}
	\caption{Performance comparison.}
	\label{table:performance}	
\end{table}

\begin{figure}[t]
	\centering
	\includegraphics[scale=0.6]{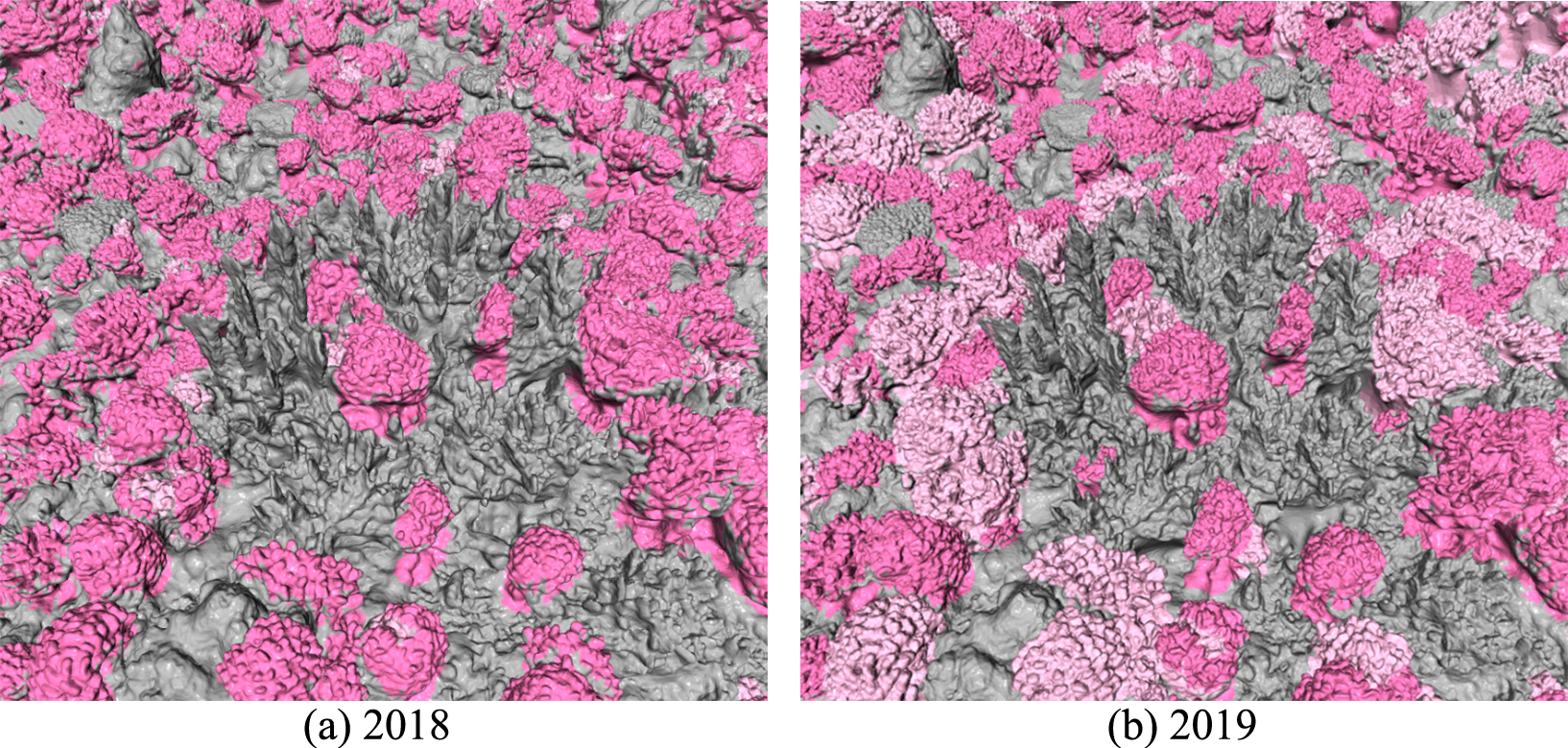}
	\caption{Visualization of the coral mesh models textured with masks from semantic segmentation.}
	\label{fig:7}
\end{figure}

Displaying coral reefs with complex spatial structures in 2D images is neither intuitive nor discoverable in their rich detail, so it becomes natural to consider about 3D visualization. We project segmentation masks from 2018 and 2019 in the study area obtained from the trained neural model above to the corresponding mesh models, as shown in Figure \ref{fig:7}. From the models in Figure \ref{fig:7}, we can see the coral reef growth variations clearly. This will help ecologists a lot to study corals in a three-dimensional, understandable and intelligent way. In Figure \ref{fig:7}, the dark pink labels are live \textit{Pocillopora} corals, the light pink labels are dead \textit{Pocillopora} corals, and the others are the background segmented by us. We can find a lot of bleaching or dying of \textit{Pocillopora} corals from 2018 to 2019. \textit{Pocillopora} corals are globally representative and are the first to be affected by the ongoing heatwaves of the oceans. The coral data studied in this paper were collected in late August 2018 and 2019 through underwater remote sensing technology, respectively. According to scientists at UC Santa Barbara, the heat wave of Moorea Island, which started in December 2018 and lasted until May 2019, was one of the strongest marine heat waves they have observed in the past 30 years~\cite{harrison2021double49}. They report that the prolonged heatwave, exceeding 29 degrees Celsius, caused about half of the \textit{Pocillopora} corals to bleach or die. This is in good agreement with the coral variation data extracted in this paper, showing the value of our method in processing coral observation data archiving and quantitative analysis.

\begin{figure}[t]
	\centering
	\includegraphics[scale=0.4]{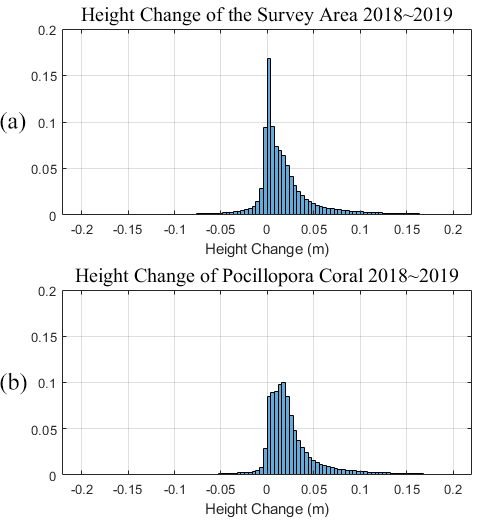}
	\caption{The frequency histogram of height changes derived from 2018 and 2019.}
	\label{fig:8}
\end{figure}

\begin{figure}[t]
	\centering
	\includegraphics[scale=0.72]{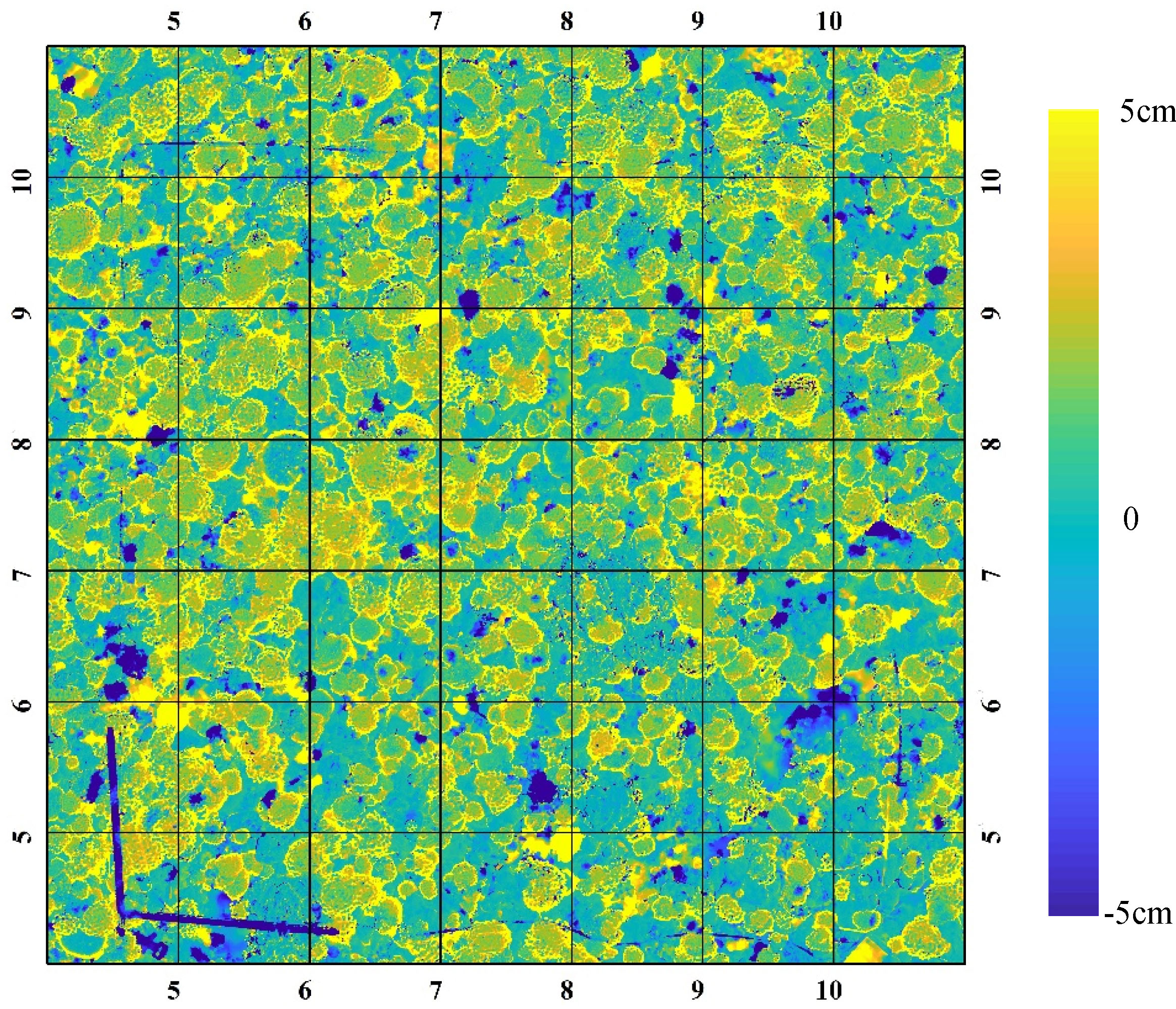}
	\caption{Changes in topographic height measured between August 2018 and August 2019. The size of each grid is 1m×1m.}
	\label{fig:9}
\end{figure}

\begin{figure}[t]
	\centering
	\includegraphics[scale=0.28]{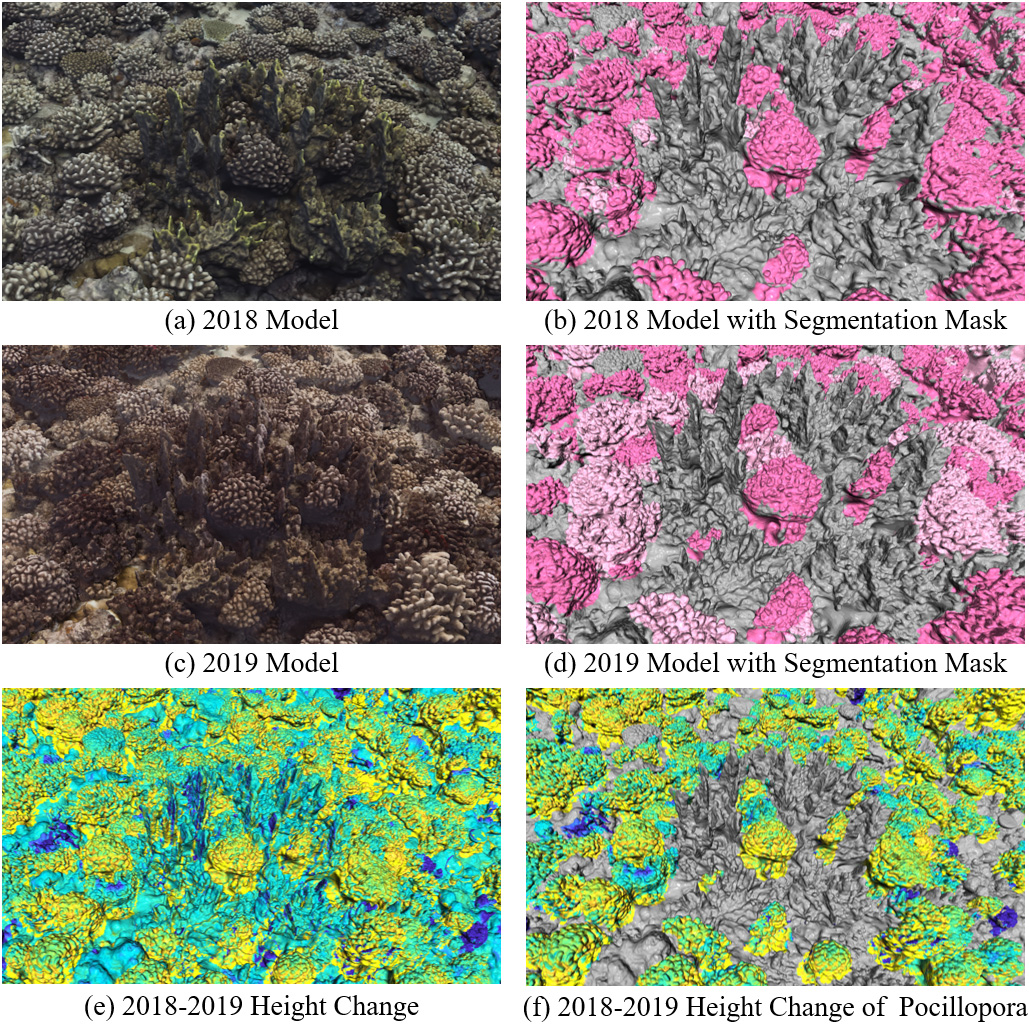}
	\caption{The height changes between 2018 and 2019 in 3D models.}
	\label{fig:10}
\end{figure}

\subsection{Height changes}

Based on the obtained high-resolution DSMs from different years, the height changes of coral reefs can be calculated. We subtract DSMs from 2018 and 2019 and make statistics of height changes at first step. The frequency histogram of height changes of the survey area is shown in Figure \ref{fig:8}(a). It apparently shows a skewed distribution, so the median appears to be a better choice than the mean for seeking a single value that reflects the location of most observations~\cite{rousselet2019reaction48}. The median of the whole survey area is calculated to be 7.7mm.

To learn the spatial distribution of height changes, we make a height change map of the survey area from DSMs at different times, where all the height changes whose absolute value is more than 50mm is truncated, as shown in Figure \ref{fig:9}. The more yellow the color is, the more height increases, and the more blue it is, the more height decreases. The nearly-right-angle dark blue descending area in the lower left corner of Figure \ref{fig:9} is caused by manually placed rulers, which exists in 2018 but not in 2019. Most areas are cyan which means there is little height change. There is only a small amount of growth at the edge of corals, and some corals even disappeared (dark blue area in Figure \ref{fig:9}). It is thus clear that corals are not growing well under the influence of the heat wave of Moorea Island. With the mask generated by MMCS-Net, we can focus on the height changes of \textit{Pocillopora} corals. Therefore, the same region in the coral reefs of 2018 and 2019 is selected for comparison, as shown in Figure \ref{fig:10}. Figure \ref{fig:10}(a) and (c) respectively show the 3D models of coral reefs of 2018 and 2019, and Figure \ref{fig:10}(b) and (d) respectively show the models textured with coral segmentation results of 2018 and 2019. The height changes of the entire area are visualized in Figure \ref{fig:10}(e), while the height changes of \textit{Pocillopora} corals are clear to see in Figure \ref{fig:10}(f). In this way, we can perform detailed geomorphological mapping and facilitate the analysis of spatial changes of the coral reef. And the frequency histogram of height changes of \textit{Pocillopora} corals can also be specially plotted, as shown in Figure \ref{fig:8}(b). The median becomes 17.8mm which is roughly double that of 7.7mm. This indicates that although the study area was affected by heatwaves in the second half of the past year, resulting in widespread bleaching or death, \textit{Pocillopora} corals grew better than other corals, which also explained its ability to become the dominant coral species in the study area.

\subsection{Reef rugosity}

\begin{figure}[t]
	\centering
	\includegraphics[scale=0.15]{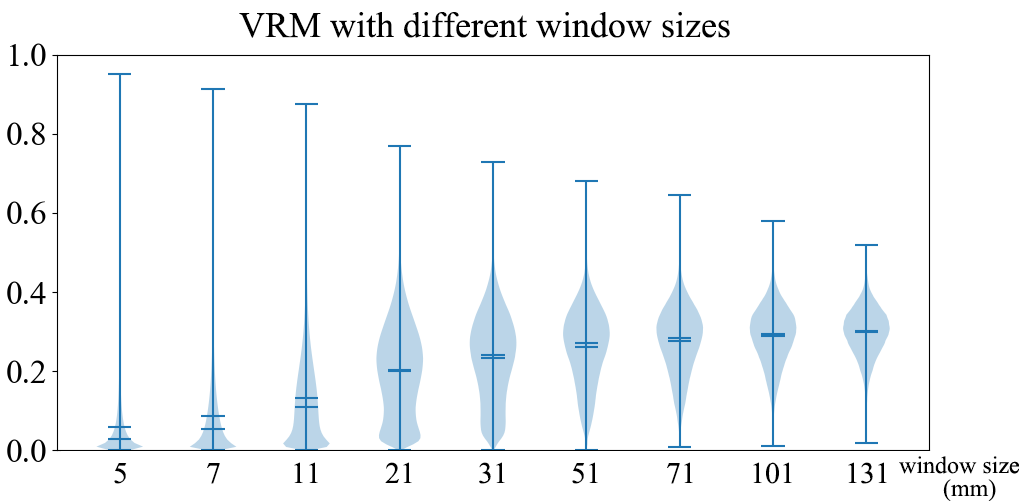}
	\caption{ Violin plots of VRM with different window sizes (5,7,11,21,31,51,71,101,131mm).}
	\label{fig:11}
\end{figure}

Fine-scale structural complexity influences biodiversity and benthic fauna abundance, and VRM can be used to reflect the structural complexity. The VRM of a coral reef is generated from DSM of 2018 with a resolution of one millimeter, and is calculated using different moving window sizes (different resolutions) to investigate a full range of scales from individual polyps to colony scales. The result is presented in Figure \ref{fig:11} using violin plot. At coarser resolutions, rugosity mainly reveals the slope of the seafloor, while finer resolutions will lead to rugosity maps that reveal a much higher granularity of variation~\cite{asner2020high13}. When the window size is smaller than 21 (21mm), most of VRM values are no more than 0.2. As the window size becomes larger, the VRM values go to 0.3 nearby. The low VRM may be caused by level sand ground, while high VRM can be the result from corals, reefs and etc. Typical benthic values are small (\textless 0.4) in natural data~\cite{asner2020high13}.

\begin{figure}[t]
	\centering
	\includegraphics[scale=0.164]{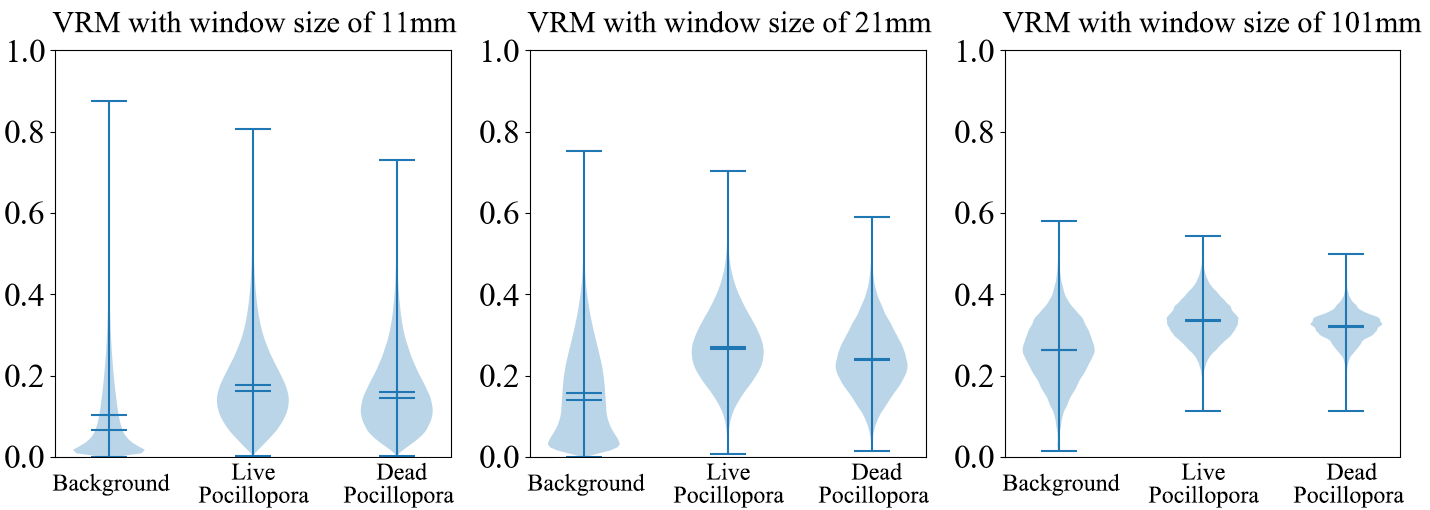}
	\caption{ Violin plots of VRM of different classes with different window sizes (11/21/101).}
	\label{fig:12}
\end{figure}

To analyze the VRM of different surface features, the VRM of these three classes (live \textit{Pocillopora}, dead \textit{Pocillopora} and background) is calculated with the mask generated from coral segmentation. From Figure \ref{fig:11}, it can be seen that there are marked differences among VRM distributions with window sizes of 11 pixel (11mm), 21 pixel (21mm) and 101 pixel (101mm). So, we calculate VRM with these window sizes, and the violin plot of VRM of three classes is shown in Figure \ref{fig:12}. The rugosity of live \textit{Pocillopora} is slightly higher than that of dead \textit{Pocillopora}, but the difference between them is not significant. This is mainly because the dead corals segmented in this paper include bleached and dead corals. Because they have not been covered by algae or other sediments for a short period of time, there is not much difference in the overall spatial complexity between them and living corals. Usually, the single skeleton of corals is white within 1 month from bleaching to death, and the structure is complete and clear; after half a year of death, it will be covered by small algae or thin layers of sediment; after 1 to 2 years of death, it will start to corrode. This also shows that monitoring the rapid changes of corals is a challenging task, which not only requires us to observe and extract information from different dimensions and perspectives, but also requires us to have the ability to monitor changes in high frequency. The method proposed in this paper has exactly such potential. Among other segmentation labels, mainly sand, corroded debris, and a small amount of coral. Their space complexity is lower than \textit{Pocillopora}, so the VRM value is also smaller. 

\section{Conclusions}

For the fine-grained understanding of coral reef growth variations, a new method using photogrammetric computer vision and semantic segmentation is proposed and applied to high-resolution coral images acquired by underwater remote sensing from Moorea in 2018 and 2019. The high-resolution (1mm) orthophotos, DSMs and mesh models of the coral reef are obtained. The coral semantic segmentation is carried out using a proposed new deep neural network that can efficiently integrate the color information from orthophotos and structure information from DSMs. On this foundation, a multi-dimensional intelligent analysis of coral reef growth from a 2D-3D perspective can be done. Moving from 2D metrics commonly employed to 3D metrics enabled by our new method can offer more realistic representations of coral reefs. For example, a vertically-oriented coral reef may contribute little to percent cover but have a large biomass relevant to metabolism, food webs and other ecological processes. Our findings illustrate that coral reefs in Moorea Island suffered from heatwave, poor growth and even widespread bleaching or mortality events. It is consistent with the fact that ongoing heatwaves have adverse effects on corals, these underwater remote sensing archived image data and extracted multi-dimensional information will help coral biologists to further analyze coral growth and recovery patterns after stressors and identify coral reef refugees. In this paper, based on underwater photogrammetry technology, a low-cost automatic method for monitoring coral reef changes is proposed, combining the latest image processing and deep learning techniques to perform detailed multi-dimensional mapping and quantitative information extraction of rapidly changing coral reefs. The novelty approach will make coral rapid change mapping routine in the near future. The discovery and monitoring of coral reef health, growth, and refugia offer an important pathway for meaningful conservation interventions, contributing to the coral reef protection under extreme climate change.

\section*{Acknowledgments}

The authors would like to thank the LIESMARS of Wuhan University and ETH Zurich for the supporting computing environment, and also thank the reviewers for their constructive suggestions. This research was funded by the National Natural Science Foundation of China (NSFC), grant number 41901407, and Wuhan University Introduction of Talent Research Start-up Foundation.
%%%%%%%%%%%%%%%%%%%%%%%%%%%%%%%%%%%%%%%%%%%%%%

{\small
	\bibliographystyle{ieee_fullname}
	\bibliography{egbib}
}

\end{document}